# Innovation Networks

## Petra Ahrweiler

## Mark Keane

**Abstract**

This paper advances a framework for modeling the component interactions between cognitive and social aspects of scientific creativity and technological innovation. Specifically, it aims to characterize Innovation Networks; those networks that involve the interplay of people, ideas and organizations to create new, technologically feasible, commercially-realizable products, processes and organizational structures. The tri-partite framework captures networks of ideas (Concept Level), people (Individual Level) and social structures (Social-Organizational Level) and the interactions between these levels. At the concept level, new ideas are the nodes that are created and linked, kept open for further investigation or closed if solved by actors at the individual or organizational levels. At the individual level, the nodes are actors linked by shared worldviews (based on shared professional, educational, experiential backgrounds) who are the builders of the concept level.  At the social-organizational level, the nodes are organizations linked by common efforts on a given project (e.g., a company-university collaboration) that by virtue of their intellectual property or rules of governance constrain the actions of individuals (at the Individual Level) or ideas (at the Concept Level).  After describing this framework and its implications we paint a number of scenarios to flesh out how it can be applied.

**Keywords**: Innovation, creativity, networks, knowledge production, society.

## 1. Introduction

The role of knowledge in modern economies is immediately obvious looking at income distributions and the share of knowledge-intensive industries in different world regions. The correlation is significant: high-tech regions match with high-income regions (cf. Krueger et al., 2004). The extensive evidence for this correlation has been monitored and documented by international and national institutions in much detail (e.g. OECD, 2009a, 2009b, European Commission 2002, 2008).

However, new knowledge does not immediately translate into technological innovation or, indeed, economic benefit. In general, scientific knowledge production simply targets a new contribution to "what is known" by following scientific standards and methods: such contributions may be inventions, discoveries, or insights and are, typically, realised in publications or a patents. Technology, according to Wikipedia, is "the making, modification, usage, and knowledge of tools, machines, techniques, crafts, systems, methods of organization, in order to solve a problem, improve a pre-existing solution to a problem, achieve a goal or perform a specific function". Not all technologies are necessarily based directly on scientific advances: Robert K. Merton (1970) describes some of the complexities in the relationship between these two realms. Innovation, however, does build on scientific creativity and the knowledge it generates, adding the dimensions of technological feasibility and commercial realisability, to target new products and processes that create economic and social value. So, fundamentally, technological innovation relies on scientific creativity.

Interestingly, both, technological innovation and scientific creativity, rest on the same social morphology: they both happen in collaborative arrangements, often characterised as networks. Collaborative knowledge production has become the dominant and most promising way to produce high-quality output in research (Bozeman and Lee, 2005), and collaborative innovation relies on this new mode of knowledge production while adding even more heterogeneous participants to the process.

The notion of 'innovation system' is quite common since the 'national innovation systems' (NIS) approach was introduced to innovation research in the 1980s. This framework (Lundvall, 1992; Nelson, 1993) again focusses on actors and their interactions embedded in a national institutional infrastructure. It concentrates on 'the systemic aspects of innovation [and of] diffusion and the relationship to social, institutional and political factors' (Fagerberg, 2003: 141). While relying on the same system concept but differentiating, elaborating and complementing the NIS approach, recent research targets sectoral systems of innovation (Malerba, 2002), technological systems, and regional innovation systems (Fornahl and Brenner, 2003).

'If anything, modern innovation theory demonstrates that a systemic perspective on innovation is necessary' (European Commission / DG Research, 2002: 25). However, when it comes down to definitions, we are left with rather vague intuitions such as:

'the dominant mode of innovation is systemic. Systemic innovation is brought about through the fission and fusion of technologies; it triggers a series of chain reactions in a total system…The interactive process of information creation and learning is crucial for systemic innovation… The characteristic trait of the new industrial society is that of continuous interactive innovation generated by the linkages across the borders of specific sectors and specific scientific disciplines' (Imai and Baba 1991: 389). Getting more precise, the definition of 'innovation system' starts to concentrate on actors and relationships. According to Beije (1998), an innovation system 'can be defined as a group of private firms, public research institutes, and several of the facilitators of innovation, who in interaction promote the creation of one or a number of technological innovations [within a framework of] institutions which promote or facilitate the diffusion or application of these technological innovations' (Beije, 1998: 256).  This perspective mainly put forward by Neo-Schumpeterian Economics has triggered extensive literature on the topic of 'innovation networks' (e.g. Koschatzky et al., 2001; Pyka and Kueppers, 2003; Pyka and Scharnhorst, 2009)  Considering these literatures, one is struck by the question of whether networks and systems the same thing?

Ahrweiler (2010) analyzes the relation between system and network concepts in innovation research and argues  that the notion of 'innovation networks' refers to the structural components of innovation (actors and their relationships) while the systemic perspective on innovation refers to a social system of dedicated communications, following the theory of Niklas Luhmann (Luhmann, 1987). This distinction takes up an earlier request to distinguish meticulously between these two levels: 'The discovery of the "organization fields", "ecological communities" or "institution networks" is not equivalent to the disclosure of the causes of innovation … At its core the point is not to unite causal innovation to networks, but rather to see in networks the organizational conditions for the dynamics of innovation' (Krohn, 1995: 31).

Innovation in networks is not easy to characterize as its processes are constantly in flux and its interactions have strong cognitive and social aspects. In innovation networks, participants from different areas/disciplines/organizations will take part from the beginning of a project and will define novel criteria for the quality control of the final product.  The need for reflection by all participants persists throughout the innovation process and, what is produced, will be socially evaluated and publically reviewable and, usually, will have to prove itself in the marketplace.  The contributions of different participants reflect varied worldviews and perspectives, in terms of which the goals and contents of each project must be negotiated.  Diverging interests struggle over the primacy of any one interpretation in defining what exactly constitutes "the innovation".

Adding even more heterogeneous actors to the process, innovation networks not only inherit but intensify the problems of collaborative knowledge production (Gibbons et al., 1994). They have, for the most part, no unanimous definition of the key problems, much less of their solutions.  This means that the characterization of innovation problems and solutions becomes the characterization of conceptual structures that are only interpretable and intelligible against the experiential background of the participants who employ the concepts. The arranging and integrating performances of negotiation networks offer the only possibility for the articulation and carrying out of singular communicative interests, according to the strategic perspectives of the individual participants, in the complex process of innovation. So, specific communicative interests are introduced as points of departure for the communication, and in any case are at the disposal of the participants throughout the entire process; for example, changing in response to altered definitions of problems, new constellations of participants, and altered strategies for solutions.

Given this novel, innovation landscape, a whole series of new questions arise. What do these processes of change and learning look like? How can the achievements of mediation demanded by the networks be produced? How does the compatibility or incompatibility of the "worldviews" of the participants come to be resolved? How can compatibility lead to integration and what strategies arise in cases of incompatibility? Indeed, we might summarise these questions as the determination of the criteria and results by which networks integrate complex models of action and communication?

## 2. The Socio-Organisational Level of Innovation

Innovation, the creation of new, technologically feasible, commercially realizable products, processes and organizational structures (Schumpeter, 1912; Fagerberg, Mowery and Nelson, 2006), emerges from the ongoing interaction processes of innovative organizations such as universities, research institutes, firms such as multi-national corporations (MNCs) and small-to-medium-sized enterprises (SMEs), government agencies, venture capitalists and others. These organizations generate and exchange knowledge, financial capital, and other resources in networks of relationships that are embedded in institutional frameworks on the local, regional, national and international level (Ahrweiler, 2010).

In recent years, network analysis of innovation networks has become a very vibrant, interdisciplinary research area. Many different aspects of innovation networks have been examined in this work: such as, studies of the binary combinations of possible actors (e.g., university-SME, university-MNC, SME-SME), of the possible links between actors (e.g., R&D alliances - Siegel, Waldman, Atwater and Link, 2003; spin-offs - Smith and Ho, 2006; licensing - Thursby and Kemp, 2002). There have also been extensive studies specifically on university-industry links (cf., Ahrweiler, Pyka and Gilbert, 2011) and on inter-firm networks (e.g., Schilling and Phelps, 2005; Porter, Whittington and Powell, 2005).

Although most of these studies have been carried out by economists and other social scientists, there have also been an increasing interest in complex networks from the physics community (in so-called, econophysics and sociophysics). "Research by physicists interested in networks has ranged widely from the cellular level, a network of chemicals connected by pathways of chemical reactions, to scientific collaboration networks, linked by co-authorships and co-citations, to the world-wide web, an immense virtual network of websites connected by hyperlinks" (Powell, Koput, Owen-Smith, 2005: 1132). Networks consisting of nodes and edges (i.e., actors and relations, or units and links) are now seen as ubiquitous, where general insights apply to their topologies, structural properties and measures (Albert and Barábasi, 2002; Newman, 2003; Halvey, Keane & Smyth, 2006; Wasserman and Faust, 1994).

The exploration of these abstract features of networks has shed new light on innovation networks. Pyka, Gilbert and Ahrweiler (2007) have explored the scale-free aspects of innovation networks (Barábasi and Albert, 1999; Watts and Strogatz, 1998; Watts, 1999). Other research has examined which network topologies are optimal for knowledge flows (Cowan, Jonard and Zimmermann, 2007; Gloor, 2006; Sorensen, Rivkin and Fleming, 2006). Another active topic is whether strong ties (i.e., friendship, contracts, face-to-face interaction) or weak ties (looser inter-personal contacts) are better for innovation (Granovetter, 1973; Uzzi, 1997; Burt, 1992, 2004; Ahuja, 2000; Walker, Kogut and Shan, 1997; Verspagen and Duysters, 2004). While many new insights have been gained from this new departure, it is really only the beginning of a fuller appreciation of the dynamics and structure of innovation networks (Koenig et al., 2009).

## 3. Why More is Needed

Many current network analyses focus on structures and states, not on what happens between the states, or on the causal mechanisms that produce states (cf. Pastor-Sartorras and Vespignani, 2007). However, what is more important and creates serious deficiencies is that most network analyses do not address the "agency dimension" of innovation networks (Ahrweiler, 2010), where innovative individuals and/or organizations move in an action space, which is co-evolving with them. The agency dimension (i.e. the possibility that actors may move intentionally in the action space) provides the processes and mechanisms for network formation and development: it is what actors do and fail to do that matters.

To address such issues we need more complex node-properties and/or more heterogeneous link-types for each node, be they people or organizations. A real-world actor moves in an action space, that consists of many dimensions; actors are permanently inventing, constructing, anticipating, changing, developing their action space, not only moving around in a given world. Actors perform different roles that require rich node descriptions concerning properties, behaviors, and states, and/or a richer link structure, which manifests what the actor does in relation to others. In short, in current network analyses the dimensions of nodes are rather limited; an organization-node in an EU R&D network simply has the relevant property of *does-EU-funded-research* with other organizations; so, other roles it may have are not captured.

Furthermore, current network analysis does not capture the particularities of knowledge generation and distribution. Network analyses deal with knowledge as a "flow substance" in a way that does not discriminate knowledge very much from what flows in other types of networks (such as, energy or information). It is the structure of the network that matters not the flow substance (i.e., knowledge). One consequence of this focus is that most network analyses address knowledge/innovation diffusion issues, but do not provide many insights on the processes resulting in the emergence of the new knowledge; a new focus that would require one to address the cognitive aspects of innovation. We try to address some of these issues in the current framework.

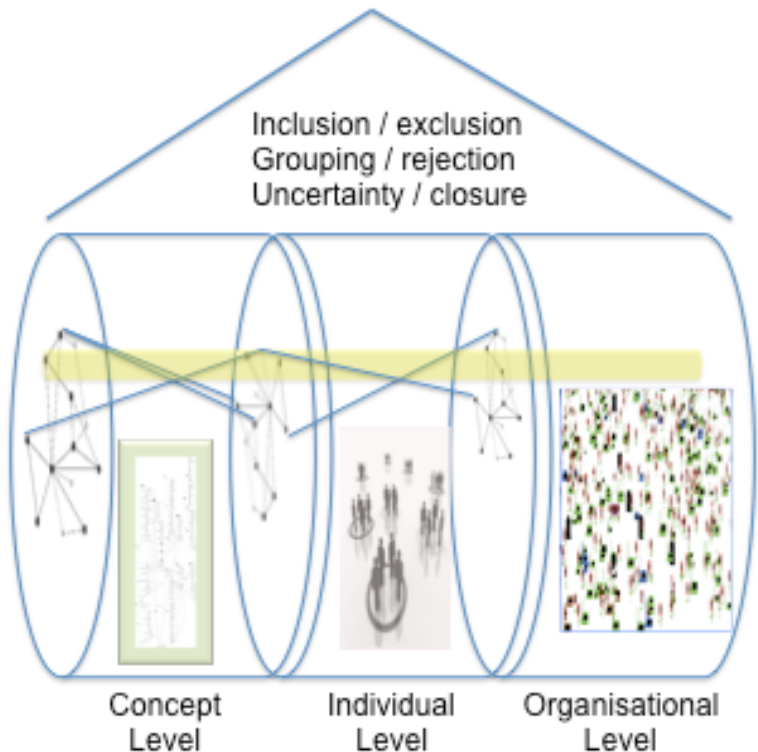

Figure 1: A framework for analysing multi-level innovation networks
(in yellow: a "slice" for analysing a real-world case navigating all levels)

## 4. The Framework

To capture the social and cognitive aspects of innovation networks it seems to be important to think in terms of, at least, three interacting layers of networks representing concepts, individuals and organizations. Keane (2010) characterizes creativity has emerging when gaps are (cognitively) opened between the World, Language (as a representational medium of describing that World) and Experience (as a conceptualization of that World). Thus, according to this view, creative individuals are people that can use language and experience to create and maintain ambiguity, to abandon previously held understandings and to balance the open and closure of the gaps that lead to creative insights.

By extension, innovation networks can be seen as social vehicles and organizational forms that somehow optimally negotiate the opening and closing of gaps in a given problem domain. Thus, networks that appear on the level of individuals forming communities of practice (Assimakopoulos, 2007; Marquis et al., 2011; Brass et al., 2004) and on the inter-organizational level as outlined above, must provide the appropriate social structures to enable the interactions between the main cognitive components of the creative universe; focusing on the requirement for ambiguity and openness in the formation processes of innovation. Inter-organizational innovation networks seem to have the task of perturbing agent-specific conceptual systems at their borders – not only with respect to specific innovation projects, but also with the role and conception of the participating organizations themselves. These networks de-construct and expose the functional fixedness and conventions of participating agents, be they individuals or collectives, enabling at the borders between organizations, gap-promoting creativity and innovation.

While previous models of innovation networks have often used ideas of interacting nodes with links between them (see above), most models have tended to be quite flat.  That is, their node-units are linked on a single plane and their interactions are all modelled along this single dimension. The current framework departs from this flat network structure by having three levels of network; the concept level, the individual level and the social-organizational level.   Each of these levels is self contained – with its own nodes, rules of linkage and grouping – but there can be interactions across levels, for instance, where a node or link is created on one level because of the presence of a given configuration at another level.   Furthermore, there is no implicit or explicit hierarchy between these levels though they are clearly interacting (hence, in Figure 1, we show them ranged horizontally to discourage an hierarchical interpretation); they are all self-contained, separate primary domains of description that do not decompose or generalise into any other level.

One of the key properties of the framework is its abstract nature.  It merely characterises unit nodes, the linkage between these nodes and their grouping but does not model the specific flavour of the link or does not model in detail the methods that create/delete links and groupings.  For example, at the concept level the nodes represent concepts and the fact that they are related in some way; it does not say whether the link means that the two concepts are "somehow associated" or "deductions from one another" or "analogically related". Neither does  the model specifiy how the association/deduction/analogy was established, it merely represents that fact that they *are* related in some way and that some cognitive process established this link. This abstract property is important, otherwise the model would have to become an omnscient theory of all cognitive and social processes that impact innovation networks (which is clearly not feasible at this stage).   Of course, in proposing various links, we may refer to such processes, but the point is that our framework does not explicitly attempt to model them.

This tri-partite framework consists of a :

- *Concept Level:* Represents the ideational structure of innovation hypotheses; the unit-nodes are concept ideas and the linkages between them show that these concepts are related (in some way) by some cognitive step taken by actors in the innovation network.

- *Individual Level:* Represents the shared worldviews (resulting from training, educational backgrounds, and so on) of the person actors in the network; the unit-nodes are people and the links are their relationships to one another established by that common worldview.

- *Social-Organizational Level*: Represents the companies, workgroups, development teams that are interacting in an innovation network; the unit nodes are heterogenous organizational groups with the links indicating that they have some formal relationship that established them as working together.

As we said above, while the linkages within these levels are important, the interactions across them are equally important. Any meaningful characterisation of the multi-facted nature of an innovation network will highlight the cross-cutting interactions between these levels to understand the dymanics and synergies that can occur.

## 4.1 Adding the Concept Level

The concept level captures the ideational structure of some collection of innovation hypotheses; here, the unit-nodes are concept ideas and the links between them indicate that certain ideas are related; where these links between concepts have been established by the cognitive or physical actions taken by actors in the innovation network. So, for example, in a lone-inventor scenario all of this level could be attributable to one person (e.g., the design for a new device) where his/her various cognitive acts serve to add new idea-nodes to the network and/or link some previously-unconnected part of the network. In a group-invention scenario this level is the result of the actions of multiple actors, such that it may well be the case that no individual actor has a view of the whole concept level (a corollary of this proposal is that there may well be regions of the concept layer that conflict one another). These cognitive acts include all those cognitive processes that change the representation of a problem; for instance, they could involve creative insights from brainstorming, newly-generated hypotheses, analogies to existing ideas, deduction/induction from known facts and so on (see e.g., Eysenck & Keane, 2005, 2010; Keane, Ledgeway & Duff, 1994; Costello & Keane, 2003). Of course, more often than not, the concept level is the result of joint contributions by multiple actors whose perspective on the problem will be shaped by their respective social and organizational backgrounds. Therefore, it obviously follows that the extent to which the group jointly appreciates and communicates to itself the scope and contents of the concept level may well be critical to its chances of success (c.f., Galison's, 1997, methaphor of 'trading zones' as a specific case).

An important property of such concept nodes is whether they can be considered to be open or closed. A closed node, is one that is considered to be solved or resolved (in some sense) and does not require further work; though it may of course be further linked to other nodes. An open node is one that remains to be solved, resolved, defined further or elaborated in some way as part of the innovation process. The proper management of the opening and closing of nodes is a critically important process to the success of any innovation network; if too many questions are left open the problem space may lack sufficient definition to be ever solved and yet if nodes are prematurely closed-off then key insights or understandings may be overlooked.

From a cognitive perspective, the opening and closing of nodes is achieved by maintaining and/or exploiting ambiguity. Keane (2010) argues, in a general account of creativity, that its essence lies in the formation/creation of comprehension gaps by exploiting the interactions between Language (spoken and written), Experience (our prior knowledge) and the World (the physical, peopled and social world out there). Creativity often arises when gaps are created between these three domains; for example, to create a new use for an existing physical object one must overcome its "functionally-fixed" current use (using a shoe heel to hammer something), one must open a gap between your prior experience of its uses to create new ones. In general, open nodes in an innovation hypothesis are the conceptual part of the problem that may be best exploited for gap-creation or gap-finding. All of the cognitive processes implicated in creativity and insight essentially act to open an existing closed node to create new understandings of the idea/problem or to bring about closure when an idea is ill-defined or poorly understood. Of course, the language-of-description adopted by the group must play a critical role in this process, as it may obscure or highlight the open-ness/closed-ness of nodes; furthermore, the language-of-description is not a given but may have to be negotiated between group-members or sub-groups with different worldviews (see also Galison, 1997).

In any given innovation episode there will be a continual updating of the set of innovation hypotheses at the concept level. Various members of the innovation network will help to create and delete nodes (should we add new ideas or reject poor ones?), to manage the status of those nodes (should we close off this part of the problem and move to defining another part of it?) and to create/delete links between nodes to integrate parts of the problem or to find a new solution (e.g., finding an analogy between one set of node-ideas and another in the problem space; c.f., Keane et al., 1994).

From this perspective, the way in which the concept level interacts with the other two levels should be clear. The person-actors at the individual level will, on the basis of their disciplinary/professional backgrounds, bring distinct perspectives to the current state of the innovation hypotheses; for example, they may be specialists in one part of the problem (e.g., design) and, therefore, add their respective constraints and key concepts to the developing understanding of the problem. From the social-organisational level particular groups (e.g., a University or a SME) may "own" part of the IP covering key concept nodes at the concept level and, therefore, be able to control acccess to these ideas.  For instance, many large IT companies defensively patent ideas to ensure that competing groups cannot use certain ideas to solve problems in their innovation space; one could imagine this as a type of *greying-out* of a whole region at the concept level, as these concepts would not be available to innovation network to use (though this exclusion could drive the development of parallel technologies/ideas to solve the problems arising).

## 4.2 Adding the Individual Level

The individual level captures the professional/educational groupings of the individuals involved in the innovation network; here, the unit-nodes are people and the links between them indicate that they share a particular worldview (based on their educational, methodological or professional background).  For example, if two people in the innovation network are electronic engineers who have worked for the same telecoms multinational for 20 years they may be linked in having a common approach to a problem, a common language-of-description for the problem and a common disciplinary understanding of that problem.  This engineering pair may have a very different approach to the problem than two social scientists in the innovation network who might focus on a different aspect of the problem space or even different aspects of the same artifact (e.g., if the group were jointly designing a new video conferencing platform, the engineers and social scientists may be focussing on the same artifact but both would be bringing very different descriptions and understandings of the device to the problem context).

This level in which individuals are linked by shared worldviews implies a rich overlap in approach. For example, professionals so linked might share a common language for describing problems (e.g., technical terms), they may also share a common methodology (e.g., using set of known statistical techniques for assessing part of the problem) and they may share a common community practice (e.g., of always partitioning a problem in a certain way). Obviously, this common worldview means that individuals in these linked groups share interpretational frameworks for understanding and exploring the innovation hypotheses of the concept level.

This individual level interacts with other two levels in diverse ways. As we mentioned above individuals with their distinct disciplinary/professional backgrounds bring different perspectives to current state of a set of innovation hypotheses. A linked group of individuals may be responsible for some whole region of the concept Level and be responsible for elaborating it, adding nodes, deleting nodes, linking notes and assessing their open/closed status. Obviously different groups at the individual level may provide complementary or competing views of the problem; for instance, to solve a particular problem it may be essential to meld insights from engineering and the social sciences from two different groups of individuals. Between the social-organisational level and the individual level there may be many complex interactions. For example, the individuals from a particular professional background may all be members of the same organization structure (e.g., company or university) or may be distributed in irregular ways across different organizations. Obviously, exactly where they sit in the organizational structure, combined with their individual background will importantly determine how they contribute to the concept level and what they bring to the project's problem. In this way, it should be clear that the interfaces between the different levels will involve many complex mappings and interactions with an emergent dynamics between the contents of the different networks.

## 5. Scenarios & Instantiations

How does this framework apply to real-world innovation network cases, for example, in knowledge-intensive industries? As mentioned before, this framework is both, rich and abstract at the same time, thus providing us with the opportunity to instantiate it in a wide variety of real-world cases. Though a full elaboration of its possible instantiations is beyond the scope of the present paper, we do provide "small lighted areas for illustration" in the current section (Figure 1, shows a yellow slice-through of all levels)[1]. Our illuminations progress in a level-wise fashion.

### 5.1 Concept Level Scenarios

Figure 2 shows an empirically-derived, concept network of the French pharmaceutical company Rhone-Poulenc on the basis of all its 1266 patents filed at the European Patent Office (EPO) from 1995 to 1999 (Pyka and Hoerlesberger, 2004). In this figure patent classes characterized by the 4-digit codes of the International Patent Classification (IPC)[2] are related to each other by links of co-citation in patent documents. The largest circle in the centre (a61k), for example, stands for "preparations for medical purposes", which is a core competence of a pharmaceutical company. The color-shaded abilities belong to the domain c12, which is biochemistry (cf. Ahrweiler, Pyka and Gilbert, 2011).

---

[1] See yellow area of Figure 1. The most obvious candidate for a "slicing exercise" would be the biotech industry, which is the best researched in terms of its networks. However, we do not know of any study empirically investigating the conceptual networks, which led to a certain innovation, while at the same time looking at individual networks targeting the same innovation, and following the relevant inter-organisational innovation networks until successful commercialisation, all in all providing information on the dynamics on all these levels and in between them.

[2] "The IPC provides a hierarchical system of language- independent symbols for the classification of patents according to the different areas of technology to which they apply. IPC Codes of patents allow the assignment of technological fields and competences with so-called concordance tables to identify industrial sectors. In this sense, the IPC Codes can be considered as coordinates of an empirical knowledge space" (Ahrweiler, Pyka and Gilbert, 2011: 222).

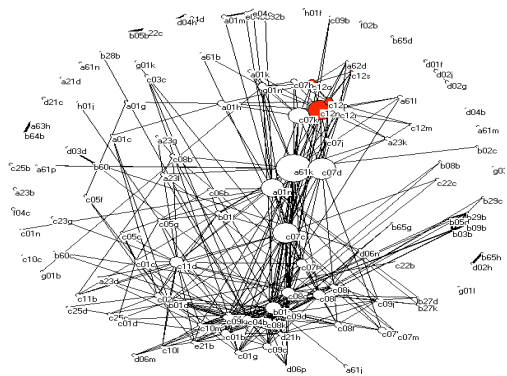

Figure 2: Knowledgebase of Rhone-Poulenc (1995-1999);
source: EPO PATSTAT (Pyka and Hoerlesberger, 2004; also see Leydesdorff et al., 2012)

This is a rather formal way to characterize concept-level innovation networks. However, one can easily imagine a more language-based version of a concept-level innovation network using linguistic-analysis techniques "as a new and innovative tool in the generation of new conceptual links in biomedical knowledge"(Vos and Rikken, 1998); such representations could be used to find "disconnected, implicit or hidden logical inference patterns in scientific literature" (Vos and Rikken, 1998: 91).

Concept-level innovation networks will not only look different due to their degree of formalisation. The above picture looks into the "brain" of a single firm and at concept interaction therein; a merger between a chemistry and a biotech firm would let us see the interaction between two different conceptual "worlds" where both would contribute their disciplinary assets for a joint purpose. In the field of so-called General Purpose Technologies (GPT) such as nanotechnology, we would see how GPT concepts would spread into various application contexts (e.g. food processing, fuel-cells, pharmaceuticals and micro-processors) and connect to the concepts found there for innovation.

In each case, we would need to look at the innovation-relevant interaction processes identified by our framework: what are the empirical manifestiations of the mechanisms for creation/deletion of nodes and links, the grouping and rejection of nodes and links and maintaining uncertainty and enabling closure in the concept network?  The knowledge map in Figure 2 shows that in the mid-1990s Rhone-Poulenc started to integrate molecular biology into their core competences as these abilities are well connected to the traditional pharmaceutical technologies (cf. Ahrweiler, Pyka and Gilbert, 2011). The framework mechanisms would have been implemented using information about the knowledge acquisition strategies of the firm at that time, and their success or failure. In this particular case, just looking at the superficial information of IPC codes in patents, the concept-level innovation would be a "designed process", namely the projection of a desired target state of the patent portfolio and its realization by management adding certain types of knowledge to the current portfolio. Inclusion/exclusion of new IPC codes would depend on access to cutting-edge knowledge, availability of resources to obtain it, and so on. Grouping/rejection to form conceptual networks with already existing IPC codes would depend on the complexity of new patents, on the potential for synergies, on issues of absorptive capacity etc. Closure would be sought when the desired state of the patent base has been reached, or when there is an alternative design choice.

Of course, the empirical manifestation of concept-level mechanisms would look different for the above examples of the inference patterns in medical databases or the interdisciplinary discussion, where it would be required to look at more difficult issues such as conceptual compatibility, reasoning, analogy, identity and other linguistic issues.

### 5.2 Individual Level Scenarios

Scenarios for the indiviudal level can be found in work on knowledge-intensive industries that cover notions of commonality, expertise and commitment. With relation to knowledge, this applies to so-called "epistemic communities", where individuals are held together by a common "recognized expertise and competence in a particular domain and an authoritative claim to policy relevant knowledge within that domain or issue-area" (Haas, 1992: 3).

Figure 3 shows the largest component of an inventor network in the Boston area for the computer devices domain, where links indicate joint patent applications (box shows high clustering). These co-inventors would have a shared understanding of what they do, can communicate easily due to shared backgrounds, and form an exclusive community of experts in their field.

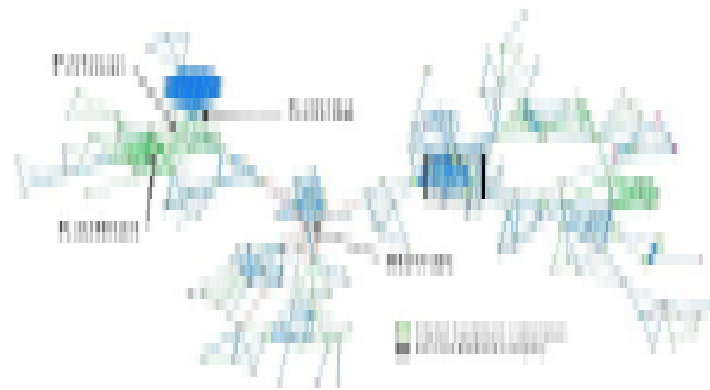

Figure 3: Inventor network of Boston 128 inventors in 1986-1990 (Fleming et al., 2007)

What we can see is that the co-patenting activity of this community consists of people working at Hewlett-Packard and various digital equipment firms in ways that criss-cross and bridge organisational boundaries.  Of course, this is always the case when two people from different companies co-patent with one another. However, organisational bridging is also a frequent and natural result from the general ability of people to move between organisations, for example as a result of entrepreneurial spin-off activity or with a usual job change.

Again, if we choose co-patenting as the activity that visibly connects people is a very formal way, it is possible to describe what is going on at the individual level of a given innovation networks. However, especially on this level, informal links are just as important. In laboratory-based innovation, it is often shared practices learnt by master-apprentice relationships, that rely on exchange of tacit knowledge, on situated learning, and on shared experience to interpret results in the right way.

In each case, we would need to look at the innovation-relevant interaction processes identified by our framework: what are the empirical manifestiations of the mechanisms for creation/deletion of nodes and links, the grouping and rejection of nodes and links and maintaining uncertainty and enabling closure in the individual network? In the above example, the answer would probably include issues such as trust, sympathy, shared language, shared expertise, former relationships and experience with collaboration, and so on.

### 5.3 Social-Organization Level Scenarios

Figure 4 shows the empirical inter-organizational network of the US biotech industry in 1997 (Powell et al., 2005). We see heterogeneous organizations involved: public research organizations such as the NIH (brown nodes), large diversified firms (yellow nodes), small and medium technology-dedicated firms (magenta nodes), venture capital firms (grey nodes) and others such as hospitals etc. These agents are involved in various collaborative arrangements such as R&D alliances (red links), commercialization (blue links), financing (green links), licensing (pink links) and others. Of course, this is again only a part of the innovation story. The market is not represented here. All these organizations would need to interact with "users", we would need to see new products and processes successfully commercialized. The relation between the network dynamics of inter-organizational innovation networks and their output, innovation performance, economic profits, and social benefits is not captured.

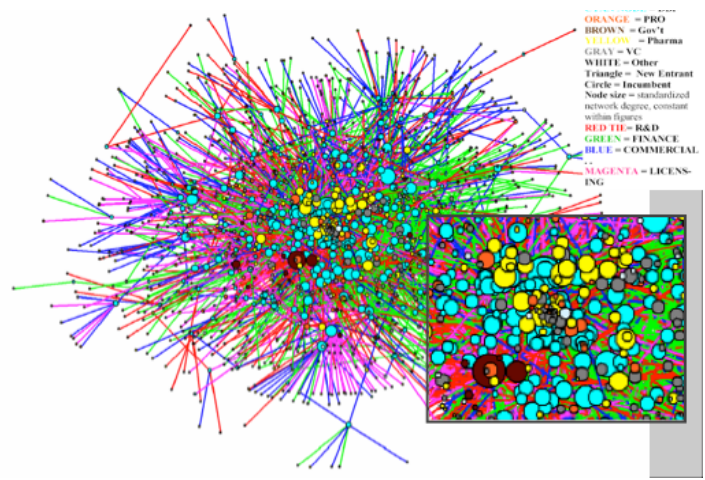

Figure 4: the US biotech industry network in 1997 (Powell et al., 2005)

Again, we would need to look at the innovation-relevant interaction processes identified by our framework: what are the empirical manifestiations of the mechanisms for creation/deletion of nodes and links, the grouping and rejection of nodes and links and maintaining uncertainty and enabling closure in the inter-organizational network?

In this case, the answer would, for example, mention issues of partner choice to allow knowledge resources to be combined, thus enabling innovation and learning that are difficult to provide by other means. It would also reflect the decreasing of risk by distributing them to network members and by accessing financial funds for the capital needed in product development; these are additional motives in these industries to decide when it is time to maintain uncertainty and when it might be time for closure.

Investigating network dynamics on the socio-organizational level relates to the aim of illuminating "how patterns of interaction emerge, take root, and transform, with ramifications for all of the participants. We develop arguments concerning how the topology of a network and the rules of attachment among its constituents guide the choice of partners and shape the trajectory of the field" (Powell et al. 2005: 5).

For Powell et al., the dynamics of innovation networks, which they define as interaction patterns between two or more organisations, can explain how fields evolve. They relate the behaviour and dynamics of the entire structure to the properties of its constituents and their interactions: individual firms learn how to collaborate with a very heterogeneous set of partners. Field evolution can be explained by the mechanisms for partner selection (Powell et al., 2005: 7f).

The strategic decisions of the networking actors and their engagement in different learning activities are responsible for the shaping of the industry. The focus is on representing the agency of innovative actors who are located in an institutional framework, on the interactions of participants and the emerging network dynamics and thus on the evolution of the industrial field (cf. Ahrweiler, Gilbert and Pyka, 2011).

Regarding the framework mechanisms of creation/deletion and grouping/rejection, it can be said that the empirical work of Powell et al. (2005) provides us with the attachment logics of the organizations in the field. These authors list strategies like "accumulative advantage" where the best-connected actors attract new/more partners, "experience-guided partner strategies" where former partners are preferred, "homophily" strategies where similar agents are chosen as partners (birds of one feather), and "multiplexity" strategies where the most different partners are attracted by one another.

## 6. Conclusion

There are many possible empirical realizations and instantiations of innovation networks. Their nodes and links are an organizational 'hardware', that arrange themselves, that can be combined, designed and composed in various settings. There is also a 'software' running on these organizational structures dealing with the availability of knowledge, finding the right partners, getting the financial resources in time, and about the smart coordination smoothing the micro-dynamics between the involved network participants and others. Which node of a given network acts as the originator, which as the transmitter, which as the enabler or receiver, can change. The actual network shape depends on participants, sectors, locations, and many other factors. The roles and processes are necessary, the actors and structures can vary.

From a scientific perspective, one of the immediate tasks in applying this framework would be to 'locate' specific studies and mathematical models from the innovation literature, as instantiations of some or all of the levels of this framework. Such a place of previous work would allow us to identify the common characteristics of these levels, their interactions and the range of processes that give rise to their creations (e.g., cognitive processes, corporate dynamics). Such a plotting of the field would also give us a good sense of which regions of the innovation network space are well- or poorly-covered; helping us to identify previously-overlooked aspects of these phenomena.

From a more applied perspective, this framework could be used to support policymakers and managers. Policymakers and managers of firms, universities and other participating organizations want to find out as much as possible about the structures and processes responsible for innovation. The managers want to know how to position their organization optimally in these networks, while the policymakers are concerned with the bird´s eye perspective on the well-being and competitiveness of the overall network on the different policy levels. Those practitioners turn to science for insights into the mechanisms and processes producing these network structures and for guidance how to optimize their performance.

# References

Ahrweiler, P., Pyka, A., and Gilbert, N. (2011) 'A New Model for University-Industry Links in Knowledge-Based Economies'. *Journal of Product Innovation Management*, 28: 218-235.

Ahrweiler, P., Gilbert, N. and Pyka, A. (2011) 'Agency and Structure. A social Simulation of knowledge-intensive Industries'. *Computational & Mathematical Organization Theory*, 17 (1):59-76.

Ahrweiler, P. (ed.) (2010) *Innovation in complex social systems.* London: Routledge.

Ahuja, G. (2000) 'Collaboration Networks, Structural Holes, and Innovation', *Administrative Science Quarterly,* 45: 425-55.

Albert, R. and Barabási, A.-L. (2002) 'Statistical Mechanics of complex Networks', *Reviews of Modern Physics T4,* 1: 47-97.

Assimakopoulos, D. (2007) *Technological communities and networks: triggers and drivers for innovation*. London and New York: Routledge.

Barabási, A.-L. and Albert, R. (1999) 'Emergence of Scaling in Random Networks', *Science* 286: 509-12.

Beije, P. (1998) *Technological Change in the Modern Economy: Basic Topics and New Developments*, Cheltenham: Elgar Publishers.

Bozeman, B. and Lee, S. (2005) 'The Impact of Research Collaboration on Scientific Productivity', *Social Science Studies* 35(5), pp. 673-702.

Brass, D. J., Galaskiewicz, J., Greve, H. R., and Wenpin T. (2004) 'Taking Stock of Networks and Organizations: a Multilevel Perspective', *Academy of Management Journal***,** *47*: 795-817.

Burt, R. S. (2004) 'Structural Holes and Good Ideas', *American Journal of Sociology,* 110: 349-99.

Burt, R. S. (1992) *Structural holes*. Cambridge, MA: Harvard University Press.

Chesbrough, H. (2003) *Open innovation: The new imperative for creating and profiting from technology*. Boston: Harvard Business School Press

Commission of the European Communities (2002) 'Benchmarking national RTD Policies: first Results', *Commission Staff Working Paper*, Brussels SEC (2002) 129.

Costello, F. J. and Keane, M. T. (2000) Efficient creativity: Constraint guided conceptual combination. Cognitive Science 24: 299-349.

Cowan, R., Jonard, N. and Zimmermann, J.-B. (2007) 'Bilateral Collaboration and the Emergence of Innovation Networks', *Management Science* 53: 1051-67.

European Commission / DG Research (2002) *Benchmarking national Research Policies: The Impact of RTD on Competitiveness and Employment (IRCSE)*. Brussels.

European Commission / EuroStat (2008) *Science, Technology and Innovation in Europe*, Brussels.

Eysenck, M.W. and Keane, M.T. (2005). *Cognitive psychology: A student's handbook*. (5th edition). London: Taylor Francis.

Eysenck, M.W. and Keane, M.T. (2010). *Cognitive psychology: A student's handbook*. (6th edition). London: Taylor Francis.

Fagerberg, J. (2003) 'Schumpeter and the Revival of Evolutionary Economics: an Appraisal of the Literature', *Journal of Evolutionary Econom*ics 13, pp. 125-59.

Fagerberg, J., Mowery, D. and Nelson, R.R. (2006) *The oxford handbook of innovation*, Oxford: Oxford University Press.

Fleming, L. and Frenken, K. (2007) 'The Evolution of Inventor Networks in the Silicon Valley and Boston Regions', *Advances in Complex Systems* 10, no. 1

Fornahl, F. and Brenner, T. (2003) *Cooperation, Networks and Institutions in Regional Innovation Systems*, Cheltenham, UK: Elgar Publishers.

Galison, P. (1997). *Image & logic: A material culture of microphysics*. Chicago: The University of Chicago Press.

Gibbons, M., Limoges, C., Nowotny, H., Schwartzman, S., Scott, P. and Trow, M. (1994) *The new production of knowledge. The dynamics of science and research in contemporary societies*, London: Sage.

Gloor, P. (2006) *Swarm creativity : Competitive advantage through collaborative innovation networks*. Oxford: Oxford University Press.

Granovetter, M. (1973) 'The Strength of Weak Ties', *American Journal of Sociology* 78: 1360-80.

Haas, P. (1992) 'Introduction: Epistemic communities and international policy coordination', *International Organization* 46: 1–35.

Halvey, M., Keane, M.T. and Smyth, B. (2006) 'Mobile web surfing is the same as Web surfing', *Communications of the ACM,* 49: 76-81.

Imai, K. and Baba, Y. (1991) 'Systemic Innovation and Cross-Border Networks, transcending Markets and Hierachies to create a new techno-economic System', *Technology and Productivity: the Challenge for Economic Policy*, Paris: OECD.

Keane, M.T. (2010) 'Creativity: A gap analysis', In *Proceedings of International Conference on Cognitive, Experience, & Creativity*. IIT Gandhinagar, India.

Keane, M.T., Ledgeway, T. and Duff, S. (1994) Constraints on analogical mapping: A comparison of three models. *Cognitive Science, 18*: 387-438.

Koenig, M. D., S. Battiston, and F. Schweitzer (2009) Modeling Evolving Innovation Networks, in: A. Pyka and A. Scharnhorst(eds.), *Innovation Networks - New Approaches in Modeling and Analyzing* Springer: Berlin, pp. 187-267.

Koschatzky, K., Kulicke, M., and A. Zenker (2001) (eds.): Innovation Networks – Concepts and Challenges in the European Perspective. Heidelberg /New York: Springer.

Krohn, W.(1995) *Die Innovationschancen partizipatorischer Technikgestaltung und diskursiver Konfliktregelung*, IWT-Paper 9/95, Bielefeld. The Innovation Chances of participatory Technology Shaping and discursive Conflict Resolution (own translation where cited).

Krueger, J., Cantner, U., Ebersberger, B., Hanusch, H. and Pyka, A. (2004) 'Twin Peaks in National Income: Parametric and Nonparametric Estimates'*, Revue Économique* 55: 1127-44.

Leydesdorff, L., D. Rotolo, and W. de Nooy (2012) Innovation as a Nonlinear Process, the Scientometric Perspective, and the Specification of an "Innovation Opportunities Explorer". Digital Libraries, http://arxiv.org/abs/1202.6235.

Luhmann, N. (1987) *Soziale Systeme. Grundriss einer allgemeinen Theorie*, Frankfurt: Suhrkamp. Social Systems. Foundation of a general Theory (own translation where cited).

Lundvall, B.-Å. (ed.) (1992) *National Innovation Systems: Towards a Theory of Innovation and Interactive Learning*, London: Pinter.

Malerba, F. (2002) 'Sectoral Systems of Innovation and Production', *Research Policy* 31, pp. 247-64.

Marquis, C., Lounsbury, M., and Greenwood, R. (Eds.), (2011) 'Communities and organizations', *Research in the Sociology of Organizations, 33:* 1-363.

Merton, R.K. (1970*) Science, technology and society in seventeenth-century England.* Harper & Row: New York.

Nelson, R. R. (ed.) (1993) *National Innovation Systems: A Comparative Analysis*, Oxford: Oxford University Press.

Newman, M. (2003) 'The Structure and Function of complex Networks', *SIAM Review* 45: 167-256.

OECD (2009a) *Policy Responses to the economic Crisis: Investing in Innovation for Long-Term Growth*, June 2009, Paris.

OECD (2009b) *Interim Report on the OECD Innovation Strategy*, June 2009, Paris.

Pastor-Satorras, R. and A. Vespignani (2007) Evolution and Structure of the Internet. A Statistical Physics Approach. Cambridge: Cambridge University Press.

Porter, K.A., Bunker Whittington, K.C. and Powell, W.W. (2005) 'The institutional Embeddedness of High-Tech Regions: Relational Foundations of the Boston Biotechnology Community', in: S. Breschi and F. Malerba (eds) *Clusters, networks, and innovation*, Oxford, UK: Oxford University Press.

Powell, W.W., White, D.R., Koput, K.W. and Owen-Smith, J. (2005) 'Network Dynamics and Field Evolution: The Growth of Inter-organizational Collaboration in the Life Sciences', *American Journal of Sociology* 110: 1132-205.

Pyka, A., N. Gilbert and P. Ahrweiler (2007) Simulating knowledge generation and distribution processes in innovation collaborations and networks. Cybernetics and Systems 38: 667-693.

Pyka, A. and M. Hörlesberger (2004) A neurological view on the technological brain of a firm, Austrian Research Centers, mimeo.

Pyka, A. and Kueppers, G. (eds) (2003) *Innovation Networks – Theory and Practice*, Cheltenham, UK: Elgar Publishers.

Pyka, A. and A. Scharnhorst (eds) *Innovation Networks. New Approaches in Modeling and Analyzing*, Berlin/New York: Springer.

Schilling, M.A. and Phelps, C.C. (2005) 'Interfirm Collaboration Networks: the Impact of Small World Connectivity on Firm Innovation', *Management Science* 53: 1113-26.

Schumpeter, J. (1912) *The theory of economic development*, Oxford: Oxford University Press.

Siegel, D. S., Waldman, D., Atwater, L. and Link, A.N. (2003) 'Commercial Knowledge Transfers from Universities to Firms: improving the Effectiveness of University-Industry Collaboration', *Journal of High Technology Management Research* 14, pp. 111-33.

Smith, H.L. and Ho, K. (2006). 'Measuring the Performance of Oxford University,

Oxford Brookes University and the Government Laboratories' Spin-Off Companies', *Research Policy* 35:1554-68.

Sorenson, O., Rivkin, J. and Fleming, L. (2006) 'Complexity, Networks and Knowledge Flow', *Research Policy* 35: 994-1017.

Thursby, J. and Kemp, S. (2002) 'Growth and productive Efficiency of University Intellectual Property Licensing', *Research Policy* 31: 109-24.

Uzzi, B. (1997) 'Social Structure and Competition in Inter-firm Networks: The Paradox of Embeddedness', *Administrative Science Quarterly* 42: 35-67.

Verspagen, B. and Duysters, G. (2004) 'The Small Worlds of strategic Technology Alliances',*Technovation* 24: 563-71.

Vos, R. and F. Rikken (1998): Connecting dis-connected structures: The modeling of scientific discovery in medical literature databases. In: Ahrweiler, P. and N. Gilbert (eds.) (1998): Computer Simulations in Science and Technology Studies. New York: Springer, 91-101.

Walker, G., Kogut, B. and Shan, W. (1997) 'Social Capital, Structural Holes and the Formation of an Industry Network', *Organization Science* 8: 108-25.

Wasserman, S. and Faust, K. (1994) *Social Network Analysis: Methods and Applications*, Cambridge: Cambridge University Press.

Watts, D. and Strogatz, S. (1998) 'Collective Dynamics of "Small-World" Networks', *Nature* 393: 440-42.

Watts, D. (1999) *Small Worlds*, Princeton: Princeton University Press.